\documentclass[letterpaper, 10 pt, conference]{ieeeconf}  

\IEEEoverridecommandlockouts                              

\title{\LARGE \bf
Natural Language Specification of Reinforcement Learning Policies through Differentiable Decision Trees 
}

\author{
\authorblockN{Pradyumna Tambwekar}\thanks{
All authors were affiliated with the School of Interactive Computing, Georgia Institute of Technology, Atlanta, GA, 30332, while conducting this research}
\authorblockA{
pradyumna.tambwekar@gatech.edu}
\and
\authorblockN{Andrew Silva}
\authorblockA{
andrew.silva@gatech.edu}
\and
\authorblockN{Nakul Gopalan}
\authorblockA{
ngopalan@gatech.edu}
\and
\authorblockN{Matthew Gombolay}
\authorblockA{
matthew.gombolay@cc.gatech.edu}
}

\usepackage{amsmath,amsfonts}
\usepackage{algorithmic}
\usepackage{algorithm}
\usepackage{array}
\usepackage[caption=false,font=normalsize,labelfont=sf,textfont=sf]{subfig}
\usepackage{textcomp}
\usepackage{stfloats}
\usepackage{url}
\usepackage{verbatim}
\usepackage{amssymb}
\usepackage{graphicx}
\usepackage{fancyhdr}
\usepackage{booktabs}       

\graphicspath{{fig/}}

\usepackage{xcolor}
\let\labelindent\relax
\usepackage{enumitem}
\usepackage{cite}

\begin{document}
\fancyhf{}

\renewcommand{\headrulewidth}{0pt}

\fancyfoot[c]{}

\fancypagestyle{FirstPage}{

\lfoot{\small \textbf{© 2023 IEEE.  Personal use of this material is permitted.  Permission from IEEE must be obtained for all other uses, in any current or future media, including reprinting/republishing this material for advertising or promotional purposes, creating new collective works, for resale or redistribution to servers or lists, or reuse of any copyrighted component of this work in other works.}}

}

\maketitle

\thispagestyle{empty}
\pagestyle{empty}

\begin{abstract}

Human-AI policy specification is a novel procedure we define in which humans can collaboratively warm-start a robot's reinforcement learning policy. 
This procedure is comprised of two steps; (1) Policy Specification, i.e. humans specifying the behavior they would like their companion robot to accomplish, and (2) Policy Optimization, i.e. the robot applying reinforcement learning to improve the initial policy. 
Existing approaches to enabling collaborative policy specification are often unintelligible black-box methods, and are not catered towards making the autonomous system accessible to a novice end-user. 
In this paper, we develop a novel collaborative framework to allow humans to initialize and interpret an autonomous agent's behavior. 
Through our framework, we enable humans to specify an initial behavior model via unstructured, natural language (NL), which we convert to lexical decision trees. 
Next, we leverage these translated specifications, to warm-start reinforcement learning and allow the agent to further optimize these potentially suboptimal policies. 
Our approach warm-starts an RL agent \color{black} by utilizing non-expert natural language specifications without incurring the additional domain exploration costs. 
We validate our approach by showing that our model is able to produce $>$80\% translation accuracy, and that policies initialized by a human can match the performance of relevant RL baselines in two domains. 
\end{abstract}

\section{INTRODUCTION}
\thispagestyle{FirstPage}
Significant progress has been made in recent years towards developing collaborative human-AI techniques that allow humans to specify a robot's desired behavior or \textit{policy}, a mapping from the state of the world to actions the autonomous agent or robot should take 
\cite{ChenPG20, chernova2014robot, shah2018bayesian}. 
However, the proliferation of such human-machine interaction hinges on the development of accessible and interpretable modes for interacting with these autonomous systems. 
Yet, while humans can provide helpful guiding behaviors for policy specifications, humans have finite cognitive capabilities~\cite{selten1990bounded} and may only be able to provide good but supobtimal policy specifications~\cite{kaiser1995obtaining}.
As such, these interactive agents need to be capable of autonomously improving upon the user's initial policy specification, e.g. via Reinforcement Learning (RL). 
In this paper, we present a novel collaborative technique that enables humans to (1) specify intelligible policies from unstructured natural language, (2) optimize these specified policies using RL.

To facilitate more accessible human-AI interactions, recent work has advocated for a shift from ``model-centered'' approaches towards deliberate design of ``human-centered'' systems focused on human-interactions~\cite{amershi2014power, ehsan2020human}. 
Collaborative policy-learning techniques require a similar degree of human-centricity to efficiently integrate autonomous systems or robots in society~\cite{sciutti2018humanizing}. 
However, many contemporary approaches to policy specification do not effectively cater to novice end-users and are unable to provide a means of interpreting an autonomous system's behavior~\cite{icarte2018using, Nilsson-RSS-18, toro2018teaching}. 

Natural language provides an accessible means of interfacing with an autonomous agent or a robot for a novice user~\cite{buchina2019natural, liu2019review}.
Prior work demonstrated increased user satisfaction when natural language is used as the interface~\cite{napier1989impact}. 
Experienced users can leverage their expertise to quickly learn how to program robots through a complex interface; however, novice users may struggle or be demotivated~\cite{hayes1985utility}. 
Therefore, we propose a methodology by which novice users can specify their desired behavior through simple and unstructured english language sentences, thus catering to the needs of a more diverse set of users.  
Policy acquisition through natural language is a well-studied area of research in recent years. 
Prior work can be broadly divided into either symbolic or connectionist approaches~\cite{smolensky1987connectionist}. 
However, given the lack of interpretability in connectionist methods~\cite{blukis2019learning, misra2018mapping, andreas2017modular} and the lack of fine-tuning in symbolic methods~\cite{arumugam2017accuratelyRSS, boteanu2017robot, gopalan18}, it is difficult to specify policies that are comprehensible to a human trainer and can improve over time. 


\begin{figure*}[t]
\centering
\includegraphics[width=0.89\linewidth]{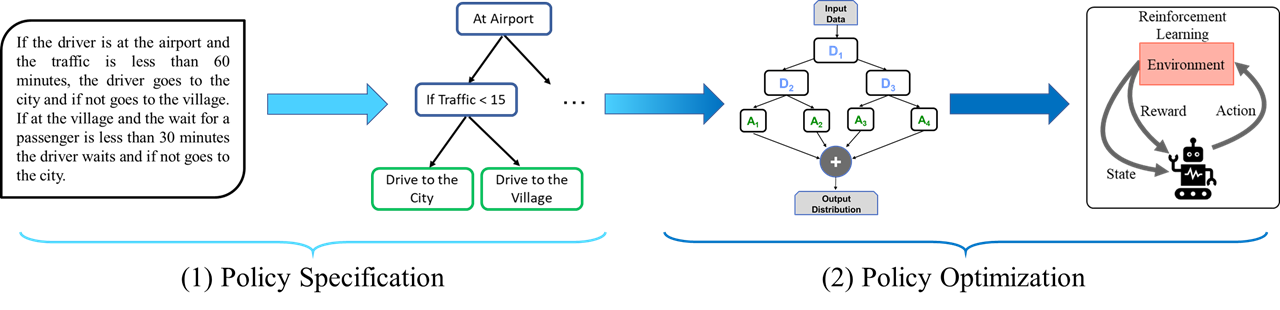}
\caption{In our Human-AI collaborative policy synthesis approach, we first convert policy descriptions to lexical decision trees. Each decision tree is encoded as a differentiable decision tree to initialize the RL policy followed by proximal policy optimization to optimize the policy.}
\label{fig:seq2tree}
\vspace{-10pt}
\end{figure*}

To address these issues, we propose a Human-AI collaborative policy synthesis architecture that bridges the benefits of both connectionist and symbolic approaches. Our framework consists of a (1) novel deep learning framework, called HAN2Tree, which learns to translate user-generated language descriptions of policies to lexical decision trees (symbolic), and a (2) Differentiable Decision Tree (DDT)~\cite{suarez1999globally} model to represent and optimize the human-specified policy with policy gradients (connectionist) given a user-defined task completion signal (i.e., a reward function). Unlike standard deep learning models, DDTs can be discretized after training into intelligible decision tree policies for users to inspect~\cite{silva2019optimization}.
We empirically demonstrate that optimized policies initialized via our natural language translation technique outperforms or matches the performance of relevant baselines across two domains by utilizing language specifications from novice users. 
Our contributions are as follows: 
\begin{enumerate}
    \item We present a data collection protocol alongside the largest known dataset mapping natural language instructions from humans to lexical decision trees (400 policy specifications).
    \item We develop a novel machine learning architecture to parse natural language instructions into lexical decision trees (HAN2Tree).
    \item We show that our method outperforms relevant baselines and obtains a translation accuracies of 86.30\% and 80.38\% across two domains and demonstrate that these translated trees can successfully warm-start RL.
\end{enumerate}

\section{Related Work}
We provide an overview of research in the areas of instruction following and interpretable ML.

\textbf{Language to Policies --}
Traditional language-based policy specification involves translating natural language to a predicate-based language for planning~\cite{ kress-gazit08, matuszek12b, tellex11, thomason2019improving}. 
For example, the sentence ``move to the left'' could map to the predicate $move(a) \wedge dir(a, left)$. This process involves a high degree of feature engineering as both the grammar and the formalizing of the predicate specification require expert design.
Some deep learning-based approaches seek to alleviate the dependency on a specified grammar~\cite{arumugam2017accuratelyRSS, gopalan18}. However, these approaches still require hand-engineering in the form of expert-specified predicates. Similar approaches also consider mapping language directly into an agent’s policy~\cite{bahdanau2018learning, Blukis:18drone, blukis2019learning} or adapt these methodologies to multi-modal systems via language and image conditioned imitation learning~\cite{lynch2020grounding, stepputtis2020language}.
These approaches condition an agent's policy via a combined learned embedding of the state and language instruction. We differentiate ourselves from these approaches by providing a means of human-centered policy specification that is simulatable and accessible, while also allowing for gradient-based policy improvement. In this context, simulatability references the model's ability to be simulated by a human~\cite{belle2021principles}.  

\textbf{Interpretable ML --}
Prior work has defined interpretability as ``the ability to explain or to present in an understandable way to humans'' ~\cite{doshi2017towards}. 
One such approach is to visualize the intermediate outputs of a neural network~\cite{ simonyan2013deep,yosinski2015understanding, zahavy2016graying}. These intermediate outputs are usually in the form of post hoc visualizations of intermediate gradients or feature maps to elicit the decision making process of a deep neural network. 
However, there is ongoing debate over whether these latent representations in a high-dimensional space actually correspond to the meanings assigned to them~\cite{montavon2018methods,szegedy2013intriguing}.  
Other forms of interpretability include providing post-hoc explanations to provide a rationale for the black-box decision making process of a model ~\cite{burns2020interpreting, frosst2017distilling, juozapaitis2019explainable, sanneman2021explaining}. Popular policy explanation methods include model-based methods, such as constrastive explanations~\cite{van2018contrastive, hoffmann2019explainable}, to consolidate competing hypotheses regarding a model's decision making process, or plan-based methods, such as policy summarization~\cite{amir2019summarizing, lage2019exploring}, which provide a means of summarizing the key details of a learned policy. 
However, these post-hoc processes are often non-trivial, and the explanations generated may not represent the conceptual model of the machine. 

Recent works on interpretable RL have developed architectures which non-experts can utilize to specify and interpret RL-policies through an inherently interpretable design, e.g. decision trees~\cite{humbird2018deep, silva2020neuralencoding}. 
Such approaches function as ``white-box'' methods, wherein the explanations provided, and interpretable capabilities come from the transparent design of the architecture itself. 
However, such approaches, when applied to policy specification, require humans to manually specify all propositional rules corresponding to an entire decision tree, which may be place a high cognitive burden on end-users in complex real-world domains.  
Combining the compositional structure of decision trees with natural language initialization is more conducive to accessible policy specification.
While language may not always be preferred to initializing trees directly, it is important to cater to a variety of end-users, some of whom may prefer language. 

\textbf{Filling the Gap --} 
The existing scope of prior work lacks accessibility for non-experts or does not allow for optimization of sub-optimal specifications. 
Overcoming these limitations, our approach takes steps towards democratizing interactive-AI systems through a transparent, simulatable methodology where end-users can warm-start the agent and interpret the final policy, 
without expert knowledge or intensive training procedures to learn the domain. 

\section{Preliminaries}
\label{sec:preliminaries}

\textbf{Differentiable Decision Trees (DDT) --} Initially introduced for classification and regression~\cite{suarez1999globally} and later extended to RL \cite{silva2020neuralencoding}, DDTs are parameterized decision trees that can be optimized through backpropagation. 
In a DDT, the $n^{th}$ node of the tree,  $D_n$, contains a set of weights, $\vec{p}_n \in P$, and comparator values, 
$c_n \in C$. At each decision node, the input state, $X$, is combined with the weights and comparator values and passed through a sigmoid, $\sigma$, approximating reasoning of a decision tree, $D_{n} = \sigma [\Gamma(\vec{p}_n^T * \vec{X} - c_n)]$, where $\Gamma$ is a scaling constant which throttles the decision making threshold. 
Every leaf node, $\vec{l_i} \in L$, contains probabilities (i.e. $l_{i,a} \in [0,1]$ such that $ \sum_{a=1}^{|A|} l_{i,a} = 1$) for each output action, $a \in A$, and a path from the root of the tree to its position. The action probabilities in $\vec{l_i}$ are weighted by the path probability of reaching $\vec{l_i}$, which is determined by the output of all decision nodes in the path. The weighted probabilities from all leaves are summed to produce the final output, which is a distribution across all actions, serving as an agent's policy, $\Pi$.

\textbf{Language Modelling} -
Recurrent neural networks (RNN) are often employed to encode sequences with temporal dependencies~\cite{hochreiter1997long}, such as language. 
A specific configuration of RNNs, sequence-to-sequence networks (Seq2Seq), have been utilized to great success for language tasks like machine translation, dialogue generation, semantic parsing, etc.~\cite{cho2014learning}.
Seq2Seq networks are comprised of an encoder, which generates a fixed size representation of the input, and a decoder, which sequentially models the probability of the next word in the sequence, $p(x_i | X_{j < i}, h)$. 
Attention was proposed as a means of improving the model's capability to remember long-term dependencies~\cite{bahdanau2014neural,luong2015effective}. These approaches produce alignment scores between the words in the input with respect to each word in the output, to produce more contextually grounded predictions. 
Recent work further expanded on attention-based networks by building sequence encoders built entirely with attention, i.e. Transformers, such that the entire text input is encoded in parallel, via self-attention~\cite{vaswani2017attention}. Transformers such as BERT~\cite{devlin2018bert} and GPT~\cite{radford2019language} are now widely being deployed to solve language applications that deal with large-scale datasets. 

\textbf{Hierarchical Attention Network} - 
The Hierarchical Attention Network (HAN) is a specific RNN architecture, that was shown to be successful for encoding larger language sequences. 
The HAN consists of two RNN layers. 
The first layer takes in embeddings, $w_{it}$, for the $t^{th}$ word in the $i^{th}$ sentence in the input and applies self-attention to create a sentence embedding, $s_i$ (Equations \ref{eq:wattn_start}-\ref{eq:wattn_end}).
\par\nobreak{\parskip0pt \small \noindent
\begin{align}
    \label{eq:wattn_start}
    s_i = \sum_{j}{\alpha_{jt}h_{ij}} \qquad\\ 
    \alpha_{it} = \dfrac{\exp({h_{it}}^\intercal W_w)}{\sum_{j}{\exp({h_{ij}}^\intercal W_w)}}\\
    \label{eq:wattn_end}
    h_{it} = GRU(h_{it-1}, w_{i:k<t})
\end{align}}\indent
Here, $h_{it}$ is the hidden vector corresponding to word  $t$ for sentence $i$. $W_w$ represents a weight vector for the self-attention layer and $\alpha_{it}$ corresponds to the importance weight for the hidden state of word $t$ in sentence $i$. 
The embedding for sentence $i$ is calculated by a weighted combination of the hidden states for every word in the sentence.
The second layer of RNNs re-applies hierarchical attention (Equations \ref{eq:wattn_start}-\ref{eq:wattn_end}) on sentence embeddings to produce the final hidden vector ($H$) for the language description \color{black}

\color{black}

\color{black}
\textbf{Sequence to Tree model - }
\label{sec:Seq2Tree}
A relevant model to this paper is the sequence to tree (Seq2Tree) network~\cite{dong2016language}. 
Seq2Tree incorporates a special \textit{non-terminal} token to decode a tree structure.
Each time the model predicts a non-terminal token, the RNN cell is re-conditioned on the non-terminal token as well as the current hidden vector of the decoder, $h_d$. 
For example, if the model predicts a non-terminal $y_4$, after three tokens $<y_1, y_2, y_3>$, the RNN cell is re-conditioned on these tokens and the model begins generating tokens $<y_5, \ldots, y_n>$ using this new RNN cell. 
A recurrent neural network is trained to maximize the likelihood of predicting the target sequence $a$ ($<y_1 \ldots y_n>$), given a hidden vector from the encoder $h_e$. For a Seq2Seq network, the likelihood is $p(a|h_e) = p(y_1,\ldots,y_n|h_e)$, as all tokens are predicted sequencially. 
In the case of this Seq2Tree example, where the a \textit{non-terminal} is predicted at $y_4$, the likelihood is $p(a|h_e) = p(y_1,y_2,y_3,y_4|h_e)p(y_5 \ldots y_n | h_d)$. 
The decoding process of a Seq2Tree network terminates when all non-terminal sequences have led to an end-of-sequence token. 

\begin{figure*}[t]
\centering
\includegraphics[width=0.86\linewidth]{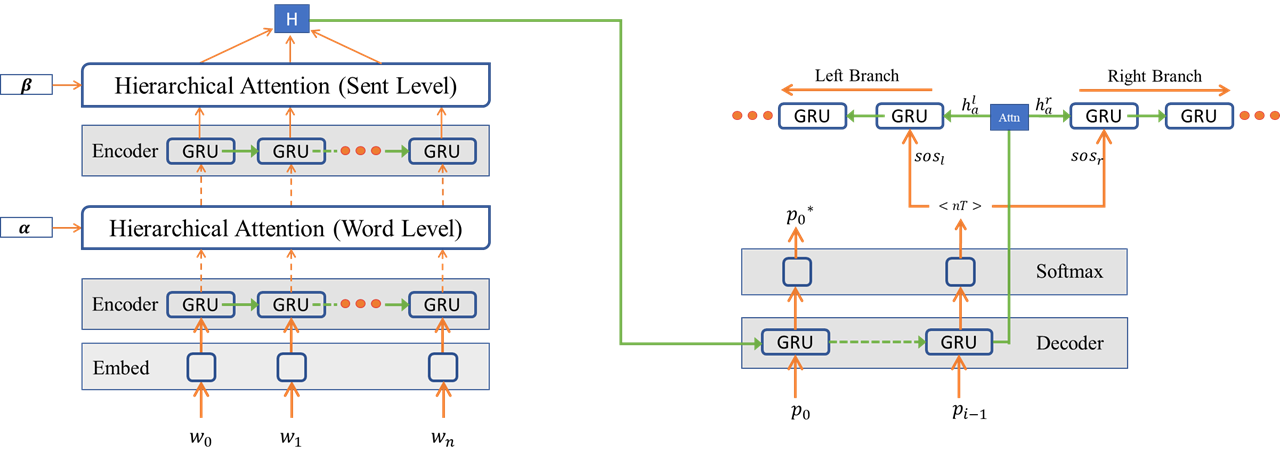}
\caption{This figure depicts our HAN2Tree architecture. The left side of the image represents the HAN encoder and the right side of the image represents the Tree decoder. Each sentence in the language description, consisting of n words ($w_0 - w_n$), is input into the HAN encoder. The encoder is comprised of two layers of GRU cells~\cite{cho2014learning}, or Gated recurrent units, which are a type of RNN-cell. The encodings for each word in the sentence are then passed into the first word-level attention layer to generate an embedding for each sentence (the attention weights for this layer are represented by $\alpha$). Next, the sentence embeddings pass through another layer of GRU cells, and a second sentence-level attention layer (the attention scores for this layer are represented by $\beta$) to generate a final encoding for the entire input ($H$). This vector is then passed to the Tree Decoder to generate the decision tree. At each step, the GRU cell predicts the next predicate in the tree ($p_i^*$). When a non-terminal token is predicted by the GRU cell, two new branches are created. Decoding stops when all branches predict end of sequence tokens or the maximum depth is reached. }
\label{fig:han2tree}
\end{figure*}

\section{Policy Specification and Optimization}
\label{sec:HAN2Tree}
\vspace{-0.8mm}
Human language can be verbose and unstructured. A successful approach to encoding human-specified policies needs to be theoretically capable of handling document-sized inputs, and identify the individual parts of the instructions pertaining to relevant components of the behavior. 
For the task of \textit{Policy Specification}, we develop an architecture, HAN2Tree, that leverages the Hierarchical Attention Network as an encoder and extends the \textit{Tree Decoder} proposed in the Seq2Tree paper (see Figure~\ref{fig:han2tree}).  

The ability of the HAN encoder to explicitly learn word-level and sentence-level dependencies allows us to better identify which parts of a disorganized language description are relevant to the policy.
A language description is encoded into a vector, $H$, and provided as an input to the \textit{Tree Decoder}.
Unlike that of Seq2Tree~\cite{dong2016language}, in our \textit{Tree Decoder}, each non-terminal prediction creates two separate branches to  facilitate a decision tree-like structure (see Figure~\ref{fig:han2tree}). At each decoding step in the \textit{Tree Decoder}, we reapply attention by learning the alignment weights of the hidden vector of the decoder with respect to the hidden vector corresponding to each sentence in the input. 
By extending the decoder from the Seq2Tree model, each branch of our decoder is able to specifically focus on information from its predecessors while using relevant information from the natural language input via attention. 


We define the target language as the set of all possible decision nodes or leaves for a given domain. 
For our experimental domains (Section \ref{sec:exp-domains}), we specify a dictionary of all possible predicates a priori.
Translating from natural language into a structured decision tree of these predicates, we are able to optimize the translated policy by encoding it as a DDT, i.e for \textit{Policy Optimization}. 
We create a mapping between each lexical predicate to a set of weights, $p_n$ and comparator values, $c_n$, as described in Section~\ref{sec:preliminaries}, to initialize the DDT through a lexical tree generated by our network.  
This intermediate decision tree representation also provides an intelligible modality for a user to visualize their specified policy, should they want to modify or re-specify the policy prior to optimization (see Figure~\ref{fig:UI_example}).  \color{black}
We can then apply proximal policy optimization~\cite{schulman2017proximal} to improve upon the initial policy specification. 
Through our approach of translating natural language into neural network structure and parameters, our HAN2Tree framework facilitates a crucial extension to prior work \cite{silva2020neuralencoding} by accommodating the needs of a larger set of end-user needs, i.e. specification via language.  
\section{Experimental Domains}

\label{sec:exp-domains}
We chose the \emph{taxi} and \emph{highway} domain, which are analogous to sub-tasks within autonomous driving. Autonomous driving is of keen interest in the robotics community~\cite{grigorescu2020survey}.

\textbf{Taxi Domain -- }
We adapt the Taxi domain~\cite{dietterich2000hierarchical} as our first domain. Our  domain consists of three locations: the airport, village and city. 
Passengers are always available at the city, however they may encounter traffic. Whereas at the village, there will be no traffic, but there may be a wait for passengers. 
The state space consists of the taxi's location, the traffic towards the city, and the village wait time. Actions consists of driving to a destination or waiting for a passenger.

\textbf{Highway Domain -- } The highway domain was initially proposed by~\cite{abbeel2004apprenticeship} for apprenticeship learning. 
The highway consists of three lanes, with the traffic all moving in the same direction~\cite{highway-env}.
The state space is comprised of the [$x$, $y$, $\dot{x}$, $\dot{y}$] vectors for the four closest cars to the agent. ($x$, $y$) corresponds to the position of the car and ($\dot{x}$, $\dot{y}$) represents the velocity of the car in the $x$ and $y$ directions.  
The agent is rewarded for safely traversing the highway at a high velocity and is given a negative reward for crashing.

\begin{figure*}[!t]
    \centering
    \includegraphics[width = 0.9\linewidth]{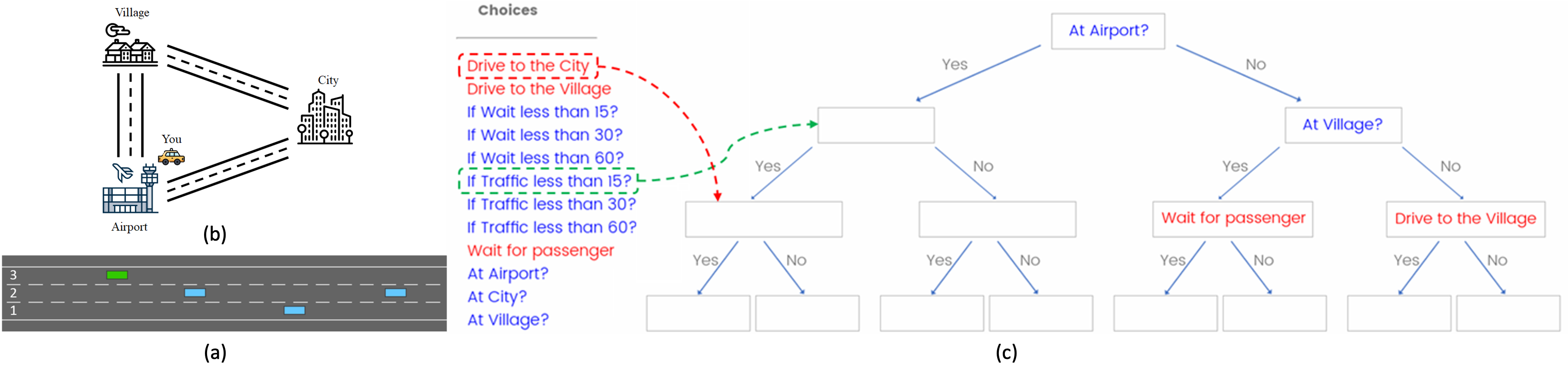}
    
    \caption{Figures (a) and (b) show the highway and taxi domains. Figure (c) provides a depiction of the interface we developed to collect decision trees corresponding to natural language descriptions of policies. Turkers could drag and drop options from a list of possible actions/decisions to generate a tree.}
    \label{fig:Turk}
\end{figure*}

    

\section{Data Collection} 

To collect our dataset, we employed Mechanical Turk to crowdsource a novel supervised learning dataset for policy specification.  
The objectives of our data collection were to: (1) Collect free-form natural language which accurately describes policies that lead to plausible behavior; and (2) Facilitate a varied set of policies specified to ensure that our model is not biased towards specific strategies.

To collect the requisite data, we built a Qualtrics survey \cite{Qualtrics} and deployed it on Mechanical Turk under a protocol approved by an Institutional Review Board (IRB). 
We did not collect demographics information for our study. Any participant between the ages of 18 - 65 from an English speaking country was eligible to participate in our study. However, we had no way of enforcing these constraints through Mechanical Turk. \color{black}
Participants did not interact with the domain; rather, users received pictures of the domain (Figure~\ref{fig:Turk}). 
Our interface contained a binary tree of depth four consisting of fifteen empty text-fields (Figure~\ref{fig:Turk}).
Participants were asked to fill in the tree using a collection of predicates to specify their desired policy. 
After creating a tree, participants were instructed to submit a natural language description of their specified tree. 
Collecting language after trees was a deliberate design choice in order to elicit language descriptions that were relevant to the participant-specified tree. 
Each submitted response was carefully evaluated according to a pre-defined rubric, and only the data points where the instructions described the policy specification were included in our final dataset. 
Our rubric included checks for whether the majority of the decisions in the tree-policy were references in the language descriptions. We also checked for whether information included in the description was copied from external sources, parts of the study itself, or was completely irrelevant to the policy specified, e.g. ``Trees are great, I liked working with trees. I enjoyed this task$\ldots$'' Minor edits were made to the submitted data based on correctness and grammar. \color{black} 
Our study collected 400 policy specifications (Text + Tree).

\subsection{Dataset Preparation}
We augmented our dataset by applying synonym replacement and back-translation on the policy descriptions. 
Back-translation is a common method in language augmentation which leverages machine translation models to translate a sentence to a different language and then translate it back to english. This serves as a means of syntactically changing a sentence while retaining the same semantic meaning. 
After augmentation, our entire dataset totalled $978$ and $998$ examples for the Taxi and Highway domains, respectively. 
Language is highly variable modality, in that two potential users could describe the same behavior in completely different ways. 
Augmenting our data adds some of this variation to our corpus and makes our model more robust to real-world language.
Our source and target vocabulary size amounted to $1283$ and $20$, respectively, for the Taxi domain and $1162$ and $20$, respectively, for the Highway domain. 
 The source vocabulary size represents the total number of words utilized in the language descriptions, and the target vocabulary represents the number of action/decision predicates among the trees in the dataset.  
We applied a 70/30 split on our augmented dataset to create our training and validation sets. 
While training, words with a frequency of less than 5 were replaced by an unknown token. \footnote{Similar language-to-structure datasets~\cite{dong2016language, jia2016data, iyer2017learning}, typically include language sequences which are almost  ${1/10}^{th}$ the length of those of our datasets, as our language inputs  describe entire policies rather than a single command (GeoQuery - 7.56, ATIS - 10.97, Scholar - 6.69, Taxi - 81.37, Highway - 92.03).
To the best of our knowledge, no pre-existing dataset is comparable to ours, in terms of size of samples or correspondence of data.}

\begin{table*}[t]
\centering

\caption{Means (standard deviations) for tree and token-wise accuracy with 5-fold cross-validation.  We report the average 5-fold accuracy across three 5-fold runs to better compare S2T and H2T, since the results of these models were very similar.}
\resizebox{0.75\linewidth}{!}{
\begin{tabular}{|c|c|c|c|c|}
    \hline
    &\multicolumn{2}{|c|}{Taxi}& \multicolumn{2}{|c|}{Highway} \\ \cline{2-5}
    & Tree Acc & Token-wise Acc & Tree Acc & Token-wise Acc \\ \hline    
    Seq2Seq & 76.54\% (0.33) &  90.35\% (0.11) &  65.11\% (0.96) &  88.12\% (0.11) \\ 
    Seq2Tree &86.04\% (0.64) & 94.83\% (0.23)  & 80.11\% (0.27) & 91.58\% (0.23)\\ 
    Seq2Tree (BERT Enhanced)&  81.84\% (1.26) & 93.07\% (0.50) & 77.17\% (0.65) & 90.38\% (0.67) \\ 
    HAN2Tree (ours)& \textbf{86.30\% (0.55)}& \textbf{95.23\% (0.12)} & \textbf{80.38\% (0.48)} & \textbf{92.83\% (0.36)}\\ \hline

    \end{tabular}
}

\label{table:results1}
\end{table*}
\vspace{-2mm}\section{Results}
In this section, we empirically validate the advantages of our approach to warm-starting policy optimization with natural language-based policy initialization.
First, we show that our approach (HAN2Tree) generates lexical decision trees from language with high accuracy, outperforming relevant baselines (Section \ref{sec:language2Policy}). 
Second, we demonstrate that we can bootstrap policies through natural language to outperform reinforcement learning baselines (Section \ref{sec:ProLoNet}). 




\begin{figure*}
    \centering
    \includegraphics[width = 0.84\linewidth]{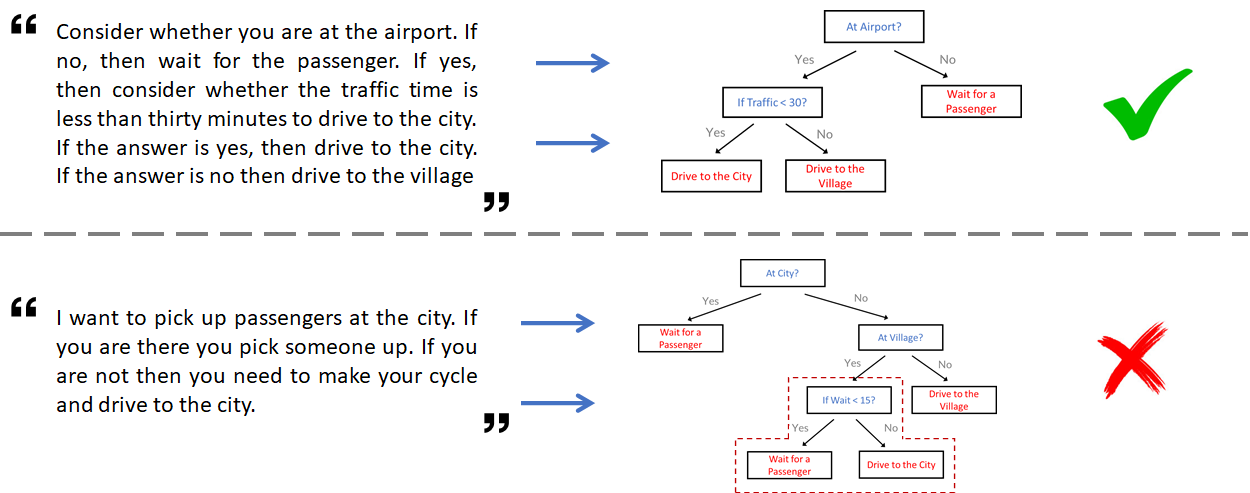}
    \caption{Two examples of trees correctly and incorrectly parsed by our system. The red box corresponds to the incorrect part of the tree predicted. The model replaced ``Drive to the City'', with a longer expression which first checked the wait time before driving to the city.}
    \label{fig:UI_example}
\end{figure*}

\subsection{Policy Specification from Language}
\label{sec:language2Policy}
We hypothesize that (1) no related network will be able to outperform our HAN2Tree architecture (2) training models from scratch is more suitable to our specialized task on a small dataset when compared to leveraging pretrained embeddings on large-scale internet corpora. 
We employ K-fold cross validation accuracy as our evaluative measure. The effectiveness of our model depends on how accurately it is able to generate decision trees from language, therefore we employed a standard classification measure utilized in prior work. \color{black}
To evaluate our Policy Specification method (HAN2Tree), we employ the following baselines: \color{black}
\begin{enumerate}
    \item Seq2Seq: We trained a Seq2Seq network with attention ~\cite{bahdanau2014neural} to generate flattened representations of trees. 
    \item Seq2Tree: An extension of the architecture presented by ~\cite{dong2016language}, with a binary Tree Decoder. 
    \item Seq2Tree (BERT enhanced): The Seq2Tree baseline augmented with pretrained BERT~\cite{devlin2018bert} embeddings. 
\end{enumerate}

We include a ``BERT enhanced'' baseline to leverage large-scale pretraining for our encoder, allowing us to see what benefits may be gained by equipping our network with this prior knowledge. 
We chose to leverage BERT as a feature encoder as prior work has shown that a fused-embedding structure is more effective means of incorporating BERT~\cite{zhu2020incorporating}. Therefore we applied the methodology used by the ELMO architecture\cite{peters2018deep} to incorporate BERT embeddings into the encoder of the Seq2Tree baseline. 
We report the 5-fold cross validation accuracy for each of our baselines (Table ~\ref{table:results1}).
Tree Accuracy is the percentage of trees that exactly matched the target tree, and the token-wise accuracy reports the per-token accuracy of the model.
Our best-performing approach achieved a mean tree accuracy of $86.30\%$ ($0.55$) in translating natural language to decision trees in the Taxi domain. We do note that the Seq2Tree approach achieves a comparable performance to HAN2Tree in both domains. However, our approach, remains a better fit for real-world applications, due to its capacity to more effectively process large text inputs. 
Despite attention, and updated RNN architectures, RNN models still struggle to retain long-term dependencies for larger input sequences.
The HAN encoder explicitly separates words and sentences in order to split up the input into meaningful segments to mitigate some of these issues. Particularly in instances with free-form language descriptions of policies, which could feasibly be comprised of $\sim$1000 words, it is important that the network has the ability to effectively segment the input and model the dependencies of each segment. \color{black}

The performance of the BERT-enhanced baseline is unable to match either the Seq2Tree or HAN2Tree approaches.
We hypothesize that due to the size of our dataset, adapting the information learnt from BERT's large-scale pretraining methodology is more challenging than learning the specific information required for this task from scratch. 
We acknowledge that the benefits of using a BERT enhanced structure, or the entire pretrained BERT transformer as an encoder, would increase in situations where data for this problems is more abundantly available.  
However, we leave this point to future work given the challenge of creating a large enough dataset for such a trend to possibly become evident.  
Sample outputs from our approach are shown in Figure~\ref{fig:UI_example}.

\subsection{Policy Optimization through RL}
\label{sec:ProLoNet}
In this section, we show that leveraging language to specify policies as DDTs is a viable method for policy learning. 
We hypothesize that (1) utilizing natural language initializations from non-expert participants does not inhibit policy learning for DDTs, and that (2) policies initialized through language, via lexical decision trees, can be sufficiently optimized to match or outperform those without language initialization.  
We compare our approach, a DDT initialized by natural language, with the following baselines:\color{black}
\begin{enumerate}
    \item NN (PPO): A neural network (NN) with three fully-connected layers, a baseline used in prior work~\cite{silva2020neuralencoding}.\color{black}
    \item Random DDT: A randomly initialized DDT initialized with eight leaves adopted from prior work~\cite{silva2020neuralencoding}.
\end{enumerate}


\begin{table}[t]
\label{tab:prolo_results}
\centering

\caption{This table displays the medians and standard errors for the initial and maximum rolling rewards for DDTs across training runs. The window-size for both the initial and maximum rolling reward was 100.  
}
\resizebox{\columnwidth}{!}{
\begin{tabular}{|c|c|c|}
    \hline
    & Taxi & Highway \\ \hline    
    NN Initial & 0.8 $\pm$ 0.43 & -3.79 $\pm$ 0.03 \\ 
    Random Initial & 3.45 $\pm$ 0.28 & -1.49 $\pm$ 0.18 \\   
    Natural Language Initial (Ours)& \textbf{4.04} $\pm$ ~\textbf{0.32} & \textbf{-1.28} $\pm$ ~\textbf{0.27} \\ \hline 
    NN Best & 8.22 $\pm$ 0.04 & -0.36 $\pm$ 0.30 \\
    Random Best & 8.26 $\pm$ 0.12 & 9.81 $\pm$ 0.43\\
    Natural Language Best (Ours) & \textbf{8.79} $\pm$ ~\textbf{0.17} & \textbf{10.00} $\pm$ ~\textbf{0.41}\\ 
    \hline

    \end{tabular}
}
\label{table:results1}
\end{table}

In Table ~\ref{tab:prolo_results}, we depict the initial and maximum rolling rewards for each baseline. For the NL baseline, we report the average of the initial and maximum rolling reward across the best 5 trees generated through natural language initializations out of a selection of 10-15 trees. The rolling reward was computed across 100 episodes. 
We find that the best NL model achieves a higher average in terms of the best performing model as well as the initial model compared to the random and FC baselines. 
The Random DDT baseline also outperforms the NN baseline, which is likely due to helpful inductive bias seen in prior work~\cite{silva2020neuralencoding}.
The average initial performance was also found to be higher for DDTs initialized by language rather than randomly initialized DDTs. 
It is interesting to note that the relative initial performance with respect to the random baseline is greater within the taxi domain (+0.6) rather than the highway domain (+0.2). 
The drop in initial performance between the taxi and highway domain, indicates that the loss in translation accuracy affects the quality of the policy initializations. 
In future work, we hope to expand our approach to a larger dataset in order to overcome this translation cost. 
However, by successfully initializing DDTs through natural language, and matching the performance of prior work, we facilitate the first such methodology, combining symbolic and connectionist  concepts, where humans can program and visualize  their desired policies through free-form natural language. \color{black}

\section{Limitations and Future Work}
In this work, we assume that specifying policies via natural language is preferable to directly specifying policies as decision trees through a graphical user interface. 
While this assumption is supported by prior work~\cite{lipton2018mythos}, it would be interesting to see if this assumption bears out in practice. 
In future work, we hope to study which modality, between natural language and decision trees, is more suitable for allowing non-experts to specify robot policies.

Another important limitation pertains to the size and scope of our dataset. 
In our data-collection protocol, participants were limited to specifying trees of depth four to reduce their cognitive load during the study. 
Based on analysis from pilot studies, trees of depth four were ascertained to be the deepest trees needed to describe the majority of behaviors possible for our domains. 
However, trees of depth four may be insufficient for other domains. 
In such cases, one would have to restructure the data-collection UI to collect variable size decision trees, but our approach would still be applicable.
With respect to the dataset, we also acknowledge the presence of an inductive bias pertaining to the method of data collection. Since we collected language after  participants submitted their trees, participants were more likely to produce language that reflected the tree structure. 
It might be possible that our approach will not perform as well on language instructions collected ``in the wild.'' However, we maintain that the building blocks presented in this approach will be crucial for future work towards developing a simulatable model for in-the-wild policies. 
\section{Conclusion}
In this paper, we bridge relevant symbolic and connectionist methods, developing a human-centered, interpretable framework for policy specification through natural language, policy improvement via reinforcement learning.
Furthermore, we showcase a data collection methodology that can be used to collect decision trees, of any size, and the natural language descriptions corresponding to these trees for any given domain.
Utilizing this protocol, we crowd-sourced the largest known corpus mapping unstructured, free-form natural language instructions to lexical decision trees. 
Our novel machine learning architecture, called HAN2Tree, was the first approach capable of generating lexical decision trees from language while outperforming a model that leveraged pretrained embeddings. 
We demonstrate the utility of using language specifications from novice users by showing that our approach outperforms or matches relevant baselines without natural language initialization.
We hope that this work promotes a greater emphasis on building accessible machine learning systems which can cater to the needs of the diverse sets of users they may interact with.

\section{Acknowledgements}
This work was supported by the Office of Naval Research under N00014-19-1-2076, the National Science Foundation under IIS-2112633 and FMRG-2229260, and a gift to the Georgia Tech Research Foundation by Konica Minolta.

\bibliographystyle{IEEEtran}
\bibliography{IEEEabrv, ref}

\end{document}